\documentclass[11pt,a4paper]{article}
\usepackage{times,latexsym}
\usepackage{url}
\usepackage[T1]{fontenc}
\usepackage[acceptedWithA]{tacl2021v1}

\usepackage[dvipsnames,svgnames,table]{xcolor}
\usepackage[disable]{todonotes}
\makeatletter
\newcommand*\iftodonotes{\if@todonotes@disabled\expandafter\@secondoftwo\else\expandafter\@firstoftwo\fi}
\makeatother
\newcommand{\noindentaftertodo}{\iftodonotes{\noindent}{}\ignorespaces}

\newcommand{\note}[4][]{{\todo[author=#2,color=#3,size=\scriptsize,fancyline,caption={},#1]{#4}}}

\newcommand{\response}[1]{\vspace{3pt}\hrule\vspace{3pt}\textbf{#1:}}
\newcommand{\jason}[2][]{\note[#1]{jason}{green!40}{#2}}

\newcommand{\cutforspace}[2][]{\note[#1]{cut for space}{gray}{#2}}

\newcommand{\Jason}[2][]{\jason[inline,#1]{#2}\noindentaftertodo}

\newlength{\extramargin}
\usepackage{calc}
\iftodonotes{\usepackage[paperwidth=\paperwidth+\extramargin*2,marginparwidth=\marginparwidth+\extramargin,width=\textwidth,height=\textheight]{geometry}}{}

\newif\ifanonymous
\anonymousfalse
\newcommand{\showornot}[2]{%
  \ifanonymous #1\else #2\fi%
}
\usepackage{times}
\usepackage{latexsym}
\usepackage[T1]{fontenc}
\usepackage[utf8]{inputenc} 
\usepackage{microtype}
\usepackage{inconsolata}
\usepackage{caption}
\usepackage{subcaption}
\usepackage{appendix}
\usepackage{enumitem}
\usepackage{graphicx}
\usepackage{amsfonts}
\usepackage{amsmath}
\usepackage[ruled,vlined]{algorithm2e}
\usepackage{booktabs}
\usepackage{multirow}
\usepackage{algpseudocode}
\usepackage{fontenc}
\usepackage{listings}
\input{listings-python.prf}

\usepackage[noabbrev,capitalize]{cleveref}

\crefname{page}{page}{pages}
\crefname{footnote}{footnote}{footnotes}
\crefname{equation}{equation}{equations}
\crefname{corollary}{Corollary}{Corollaries}
\crefname{line}{line}{lines}
\crefrangeformat{line}{lines #3#1#4--#5#2#6}
\crefname{lstlsting}{Listing}{Listings}   
\crefname{section}{\S}{\S\S}
\Crefname{section}{\S}{\S\S}
\crefformat{section}{#2\S#1#3}
\Crefformat{section}{#2\S#1#3}
\crefrangeformat{section}{\S\S#3#1#4--#5#2#6}
\Crefrangeformat{section}{\S\S#3#1#4--#5#2#6}
\crefmultiformat{section}{\S#2#1#3}{ and~\S#2#1#3}{, \S#2#1#3}{ and~\S#2#1#3}
\Crefmultiformat{section}{\S#2#1#3}{ and~\S#2#1#3}{, \S#2#1#3}{ and~\S#2#1#3}
\crefrangemultiformat{section}{\S\S#3#1#4--#5#2#6}{ and~\S\S#3#1#4--#5#2#6}{, \S\S#3#1#4--#5#2#6}{ and~\S\S#3#1#4--#5#2#6}
\Crefrangemultiformat{section}{\S\S#3#1#4--#5#2#6}{ and~\S\S#3#1#4--#5#2#6}{, \S\S#3#1#4--#5#2#6}{ and~\S\S#3#1#4--#5#2#6}

\def\t#1{\text{#1}}

\newcommand{\defn}[1]{\textbf{#1}}

\usepackage{xspace,mfirstuc,tabulary}

\newif\iftaclinstructions
\taclinstructionsfalse
\iftaclinstructions
\renewcommand{\confidential}{}
\renewcommand{\anonsubtext}{(No author info supplied here, for consistency with
TACL-submission anonymization requirements)}
\newcommand{\instr}
\fi

\iftaclpubformat

\else

\fi

\title{Accelerating Language Model Workflows with \textsc{Prompt Choreography}}

\author{TJ Bai \and Jason Eisner \\
  Johns Hopkins University \\ Baltimore, MD \\
  \texttt{\{tbai4,eisner\}@jhu.edu}}

\begin{document}

\maketitle
\begin{abstract}
Large language models are increasingly deployed in multi-agent workflows. We introduce Prompt Choreography, a framework that 
efficiently executes
LLM workflows by maintaining a dynamic, global KV cache. Each LLM call can attend to an arbitrary, reordered subset of previously encoded messages.  Parallel calls are supported.  Though caching messages' encodings sometimes gives different results from re-encoding them in a new context, we show in diverse settings that fine-tuning the LLM to work with the cache can help it mimic the original results.  
Prompt Choreography significantly reduces per-message latency (2.0--6.2$\times$ faster time-to-first-token) and achieves substantial end-to-end speedups (>2.2$\times$) in some workflows dominated by redundant computation.
\end{abstract}

\section{Introduction}

Large language models (LLMs) are increasingly deployed beyond simple prompt-response interactions in multi-step \defn{workflows} that compose many LLM calls across interconnected \defn{agents}. These workflows have driven measurable progress across diverse domains \citep{guo2024largelanguagemodelbased}.

We introduce Prompt Choreography, a framework for Transformer LLMs where every LLM call is instructed to attend over some \emph{arbitrary reordered subset} of previously encoded messages.  This mechanism frees workflow developers to break from traditional prompted autoregressive decoding, in which each decoded token attends to \emph{all} previous tokens.
Developers can strategically reuse cached Transformer key-value (KV) \defn{encodings} to reduce redundant computation, while still choosing which messages should be visible to each agent, and in what positions.

The traditional approach requires each call to the LLM to encode the entire prompt from scratch.
Yet as agents work on a problem, 
it is common to reuse input and output messages across multiple calls.  After all, each agent usually conditions on most of its own previous output and on fixed system instructions, and multiple agents usually share substantial context, such as background documents and previous inter-agent communications.  While \defn{prefix caching} strategies \citep{zheng2024sglang, ye-etal-2024-chunkattention} will reuse some message encodings for some workflows, this simple optimization must be generalized for multi-agent workflows to reap similar benefits.

\begin{figure}[t]
    \centering

    \includegraphics[width=0.48\textwidth]{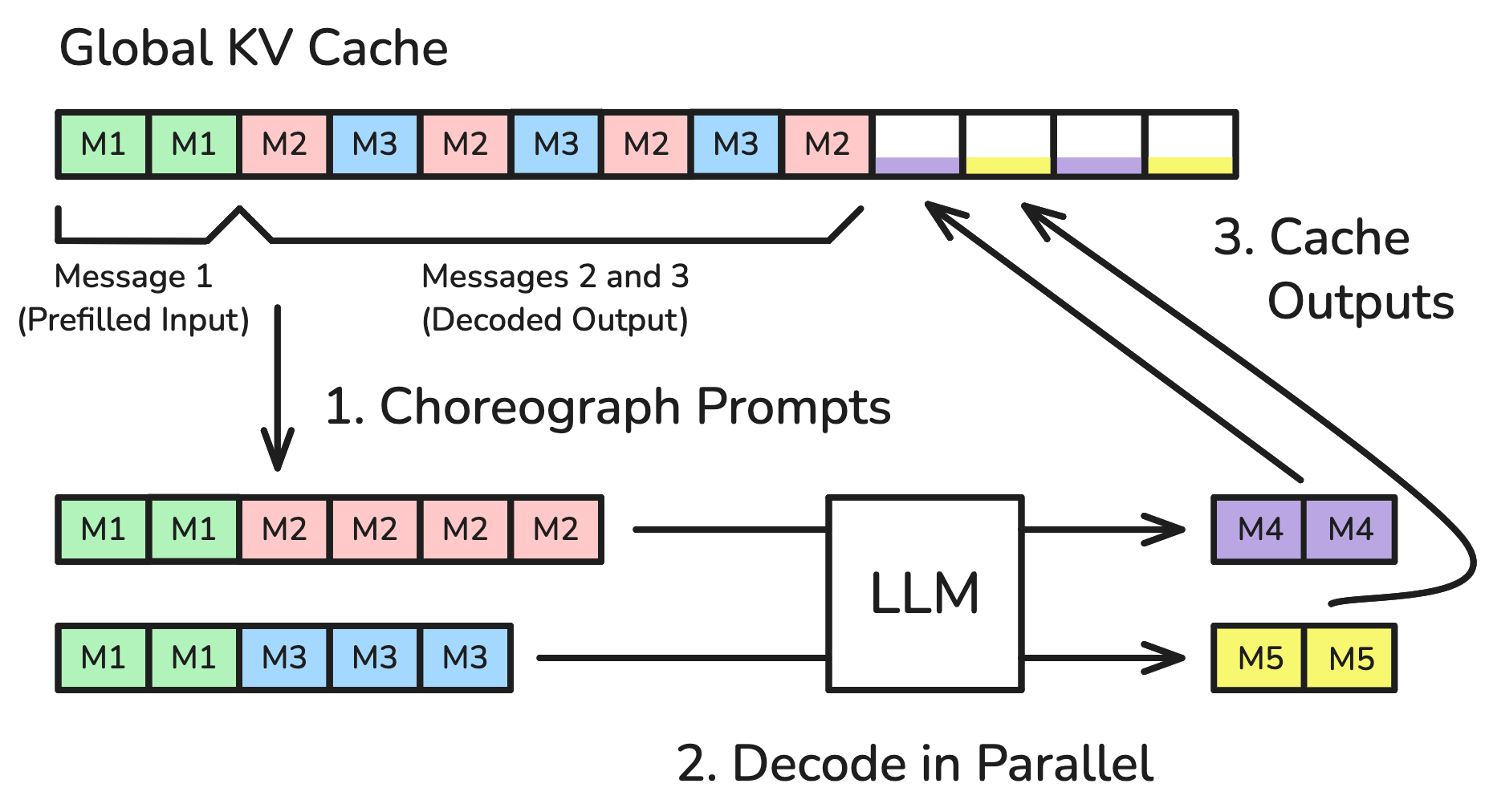}
    \Jason{the braces under the KV Cache look odd - there's a curly brace shape available in Powerpoint.}
    \caption{Prompt Choreography manages a global KV cache of messages, which is shared and extended by all participating agents.
    When assembling a prompt, an agent selects messages and may specify their start positions; this allows reordering, gaps, and overlaps (not shown).  Parallel decoding is possible whenever multiple prompts can be created from the current cache state.\looseness=-1}
    \label{fig:overview}
\end{figure}

Previous methods, particularly Prompt Cache \citep{gim2024promptcache}, partially address this by pre-computing a \emph{cache} of messages that can be selectively used at run-time, such as contextual documents. However, these approaches are generally static; messages that are dynamically generated at run-time cannot be reused. Prompt Choreography overcomes this limitation by introducing a \defn{global KV cache} that can be arbitrarily updated and accessed by all agents at run-time.

Drawing an analogy to computing architectures, traditional LLM agents are processes in a distributed memory model, where each process has a \emph{private} context window, so sharing information requires expensive copying or re-computation. Prompt Choreography instead opts for a \emph{shared} memory model \citep{lamport}, where processes access \emph{virtual} views of a dynamic global context, allowing computational states to be efficiently shared while maintaining isolation when needed.

In \cref{sec:core-idea}, we develop the core ideas behind Prompt Choreography and describe how it may be implemented and used in practice, as in our reference implementation.\footnote{\showornot{[Anonymous URL]}{\url{https://github.com/tjbai/choreo}}}  Our approach uses several novel techniques to enable fine-grained KV cache management while maintaining ease of use.  We combine a dynamic attention masking strategy (controlling \emph{which} previous messages each agent sees) with efficient position updates
(controlling \emph{where} those messages appear in the prompt) to support virtualized, parallel generation.
 This allows messages to be decoded fully in parallel over a shared KV cache while roughly maintaining appropriate logical isolation of agents. Together, these methods significantly reduce redundant computation while facilitating efficient, parallel generation.

Using Prompt Choreography does require care.  It sometimes results in different message encodings than in a standard workflow (for good or for ill).  For example, it is now possible for message 3 to attend to both messages 1 and 2, each of which was encoded into KV vectors without attention to the other.
In \cref{sec:approximate}, we discuss this kind of \defn{information blockage}, which makes messages more independent,
as well as \defn{information leakage}, where a choreography---in the name of efficiency---allows agent B indirect access to agent A's private context by reusing agent A's own encoding of an output message.
Through targeted experiments, we examine the potential adverse downstream impact of choreography.  We then show that lightweight \emph{parameter-efficient fine-tuning} effectively and efficiently mitigates these issues.

In \cref{sec:experiments}, we evaluate three representative workflows on the standard MATH benchmark \citep{MATH}. 
While choreographed workflows may underperform with an LLM that was not trained for such usage,
fine-tuning on a few hundred examples quickly regains, and sometimes exceeds, baseline performance. The fine-tuning work is then amortized by a nice run-time speedup—the resulting workflows achieve between 2.0--6.2$\times$ faster time-to-first token and consistent end-to-end speedups. Through further scaling in \defn{prefill-bound} workflows, we show that Prompt Choreography can obtain up to a 2.2$\times$ end-to-end speedup.

\begin{figure*}[t]
    \includegraphics[width=\textwidth]{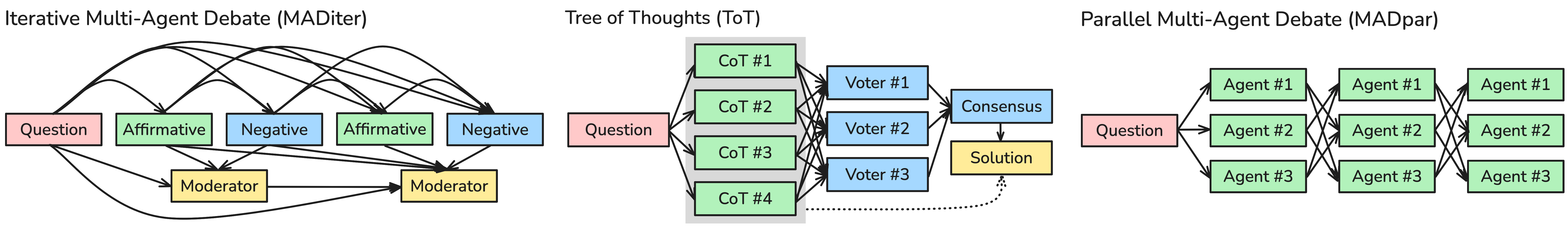}
    \caption{Workflows for the experiments of \cref{sec:experiments}.
    Full code sketches using our API appear in \cref{fig:examples-full} in \cref{app:workflow-details}.  Each box is a message, with arrows to it from its parents.  Pink boxes are prefilled.  Each non-pink box has an additional parent (not shown): a system prompt corresponding to its color.  For instance, each blue ``Voter'' in \textbf{middle} is generated with attention to instructions on how to select the best candidate.}
    \label{fig:examples}
\end{figure*}

\section{Prompt Choreography}

\subsection{Core Idea}
\label{sec:core-idea}

We extend the industry-standard \defn{Chat API} for accessing LLMs 
\cite{openai-chat-api-announcement-2024}.  The Chat API is invoked with a sequence of \defn{messages}---text strings annotated with agent roles.  Conditioned on a concatenation of these messages, the LLM generates and returns a new message.  All messages are tokenized internally.

A message typically corresponds to a turn in a dialogue, an example input or output, a document to read, or an instruction.  
Messages are natural units for caching because a message is often reused in its entirety across prompts, with essentially the same meaning each time.\jason{This is weaker than saying that its meaning is independent of other messages (context-free), which is sort of what we wrote before: "because their \emph{internal} meaning remains relatively stable even when the \emph{external} context changes."  For example, the meaning of an input may depend on  a system instruction about how to interpret its formatting, which is why we allow messages to have specified parents.\response{jason}Here's an interesting experiment: encode an input with a system instruction as its parent, and then decode an output with only the input as its parent.  Does the encoded input effectively remember the system instruction?  I could imagine that if the instruction says "Use the secret code: apple is code for banana" and the input says "Jack has 2 apples and Jill sells him 3 more apples---what does he have now?" the answer might be "5 bananas" if the decode can't see the system instruction, but "5 apples" if it can!}
This stability means that the encodings computed by one agent can often be effectively reused by another agent, eliminating redundant work in workflows that exhibit large amounts of message reuse.

Prompt Choreography maintains a global \emph{cache} of messages that are shared by all agents throughout a workflow's execution. Each message comprises not only a span of tokens, but also their corresponding Transformer KV encodings.
LLM calls add new input or output messages to the cache, conditioning their encodings on any subset of the previously cached messages.
A \emph{prompt choreography} is an arbitrary program that specifies how each call should select and arrange this subset.

Implementing this approach requires addressing three key issues:

First, retrieving cached encodings must be faster than simply recomputing them (which is already efficient with GPU parallelism). We accomplish this through memory locality, keeping the cache on the same device as the LLM and using a dynamic attention mask computed on-the-fly to control which cached encodings each new message accesses.

Second,
LLMs care about the relative position of previous tokens, so we must control where to place the selected messages.  Our position updating technique assumes a relative positional encoding scheme such as RoPE \citep{su2023roformerenhancedTransformerrotary}.  Under this scheme, the KV cache 
is translationally invariant,
so we can arbitrarily reposition messages without full recomputation.

Third,
the Chat API allows parallel sampling of multiple responses to the \emph{same}  prompt.  We extend this ability and allow multiple messages to be decoded simultaneously, each attending to a \emph{different} prompt that is choreographed from the same KV cache.  Our implementation interleaves the tokens of the decoded-in-parallel messages as it appends them to cache storage.

\subsection{Simplifying Assumptions}
\label{sec:implementation}

We will make the following practical assumptions:
\begin{enumerate}  
    \item The global KV cache fits entirely in GPU memory, allowing cached encodings to be easily attended to during inference.\footnote{\label{fn:evict}When this assumption does not hold, one could temporarily swap messages out to CPU, reduce the memory footprint through cache compression, or drop less important tokens via cache eviction. See \citet{li2025surveylargelanguagemodel}.}

    \item LLM calls are generated programmatically and fairly rapidly.  There are no pauses in the choreography---e.g., to wait for a human dialogue participant or a slow software tool to provide the next message.  Thus, executing the choreography does not selfishly lock GPU memory that may be needed by other workflows running on the same LLM server.\footnote{Again, this could be mitigated by swapping the cache out to CPU memory when the choreography is idle.}

    \item The encoding of a token does not reflect its absolute position in the prompt \cite{vaswani2023attentionneed},
but only its position relative to other tokens \cite{press2022trainshorttestlong,su2023roformerenhancedTransformerrotary}.  This lets us reposition past messages relative to the start of a new message before sequentially generating and encoding the new message's tokens.
    Our implementation assumes the currently popular RoPE scheme for relative position embeddings \cite{su2023roformerenhancedTransformerrotary}, since it is used by the LLMs we experiment with. 
\end{enumerate}

\subsection{A Prompt Choreography API}
\label{sec:api}

The standard Chat API provides a single function, $\texttt{complete}(\texttt{inputs}) \rightarrow \texttt{output}$, which autoregressively generates an output message conditioned on a ordered list of input messages. Internally, this operation can be decomposed into two phases: a parallel \defn{prefill} phase that computes encodings for all input messages at once, followed by a sequential \defn{decode} phase that produces the output message token-by-token.\jason{should we guarantee that message ids are assigned sequentially? Conceivably this would also simplify calls.}

Our API explicitly separates these phases into \texttt{prefill} and \texttt{decode} functions. Each function appends a newly encoded message to the global KV cache and returns a unique new identifier for this message for future reference.\footnote{Fancier parallel versions of these functions, which are discussed in \cref{sec:parallel}, can return \emph{multiple} new message identifiers.}  (A message's textual content can be retrieved from its identifier.)

\begin{enumerate}
    \item \texttt{\small prefill(message: str, \\
    \phantom{prefill(}parents: List[id], \\
    \phantom{prefill(}offsets: List[Optional[int]], \\
    \phantom{prefill(}new\_offset: Optional[int])} $\to$ \texttt{\small id}

    Tokenizes \texttt{message} and encodes its tokens in parallel, allowing each to attend to the preceding tokens and also to all tokens in the existing messages \texttt{parents}.  Returns an identifier for the resulting prefilled message.  For purposes of computing relative-position attention, the \texttt{parents} are repositioned to start at the respective \texttt{offsets},\footnote{Repositioning might not actually be essential unless the past messages need to be reordered.  The LLM might be robust to gaps and overlaps among messages in the prompt, either off-the-shelf
    \citep{gim2024promptcache} or after our fine-tuning (\cref{sec:ft}).\looseness=-1}\jason{
    experiment in final version to see whether repositioning is needed?
    \response{TJ} We don't have these ablations because it doesn't actually make sense in most of our settings, so I leave positions untouched during and after fine-tuning. Either we want to leave the messages in place to avoid positional bias or the structure of the workflow means the messages already appear in the ``correct'' positions. In the former case, we're probably missing an ablation study.}  and the new message is positioned at \texttt{new\_offset}. Any omitted offset defaults to the position immediately after the end of the preceding parent message.

    \item \texttt{\small decode(header: ~str, \\
    \phantom{decode(}parents: List[id], \\
    \phantom{decode(}offsets: List[Optional[int]], \\
    \phantom{decode(}new\_offset: Optional[int])} $\to$ \texttt{\small id}\jason[color=yellow]{I moved header to the start to make the two calls more similar.  TJ, can you change the codebase correspondingly?}

    Generates\footnote{Outputs may be generated using an arbitrary decoding scheme, such as temperature sampling or beam search.}
    a new message, conditioning each token and its encoding on the encodings of previous tokens and messages, with relative-position attention as before.  Returns an identifier for the new message.  The new message is constrained to start with \texttt{header}---for example, ``\texttt{Assistant:}'' for a role-based output or ``\texttt{\{}'' for a JSON output. 
\end{enumerate}

\noindent The \texttt{header} in \texttt{decode} must be non-empty,\footnote{LLMs standardly require special tokens in each message, in particular to mark the start/end of the message and specify its role. We have presented \texttt{prefill} and \texttt{decode} here as if the caller must provide these tokens, but in fact our implementation adds them transparently.} since the first unconstrained token will be generated from the top-layer encoding of the last header token.  (The list of $\geq 0$ repositioned parents may not have any obvious ``last token'' to use for this purpose.)

\subsection{Implementation: Managing the KV Cache}
\label{sec:managing}\label{sec:parallel}

Suppose the given Transformer language model \citep{vaswani2023attentionneed} 
has 
$L$ layers, each employing $h$ attention heads that consume separate keys and values in $\mathbb{R}^d$.  Then all the keys and values for a single token can be gathered into tensors $K, V \in \mathbb{R}^{L\times h\times d}$.
Caching these tensors makes it fast to attend to that token in the future.  

The global KV cache stores $K$ and $V$ for all previously \texttt{prefill}ed or \texttt{decode}d tokens.  These reside contiguously in GPU memory.\cutforspace{To optimize performance and avoid reallocations, these buffers are pre-allocated to a generous maximum size.}  We maintain\cutforspace{ a \texttt{cache\_len} pointer that is incremented to match the current number} a count of currently cached tokens, and new messages\cutforspace{As new encodings are computed during \texttt{prefill} and \texttt{decode} calls, they} are appended sequentially to the end of the currently occupied portion of the cache.  This append-only strategy is simple and fast.\footnote{See \cref{fn:evict} for potential enhancements.}

When \texttt{prefill} or \texttt{decode} appends a new message with identifier $m$ at \texttt{new\_offset} $o$, it rotates each new token's key vectors under the RoPE scheme to ``place'' the $i$\textsuperscript{th} token at logical position $j=o+i$ within the prompt.  We store the small integer pair $(m,j)$ alongside the token's key vector.  If a future API call needs to reposition this message, it modifies $j$ and the rotated key vector---doing so non-destructively during model fine-tuning (to support backpropagation through the computation graph), but destructively during inference.\footnote{If attention does not mutate key encodings to apply positional embeddings, such as in ALiBi \citep{press2022trainshorttestlong} and T5 \citep{2020t5}, then this distinction is unnecessary. \label{note:alibi}}\jason{perhaps mention that we also need a map to be able to find the parent messages: e.g., message ids are allocated as $0,1,\dots$ and message $i$ covers $[\texttt{offset}[i], \texttt{offset}[i+1])$.  Hmm, there's actually some cutforspace stuff about this above.}

\begin{figure}[t]
    \centering
    \includegraphics[width=0.48\textwidth]{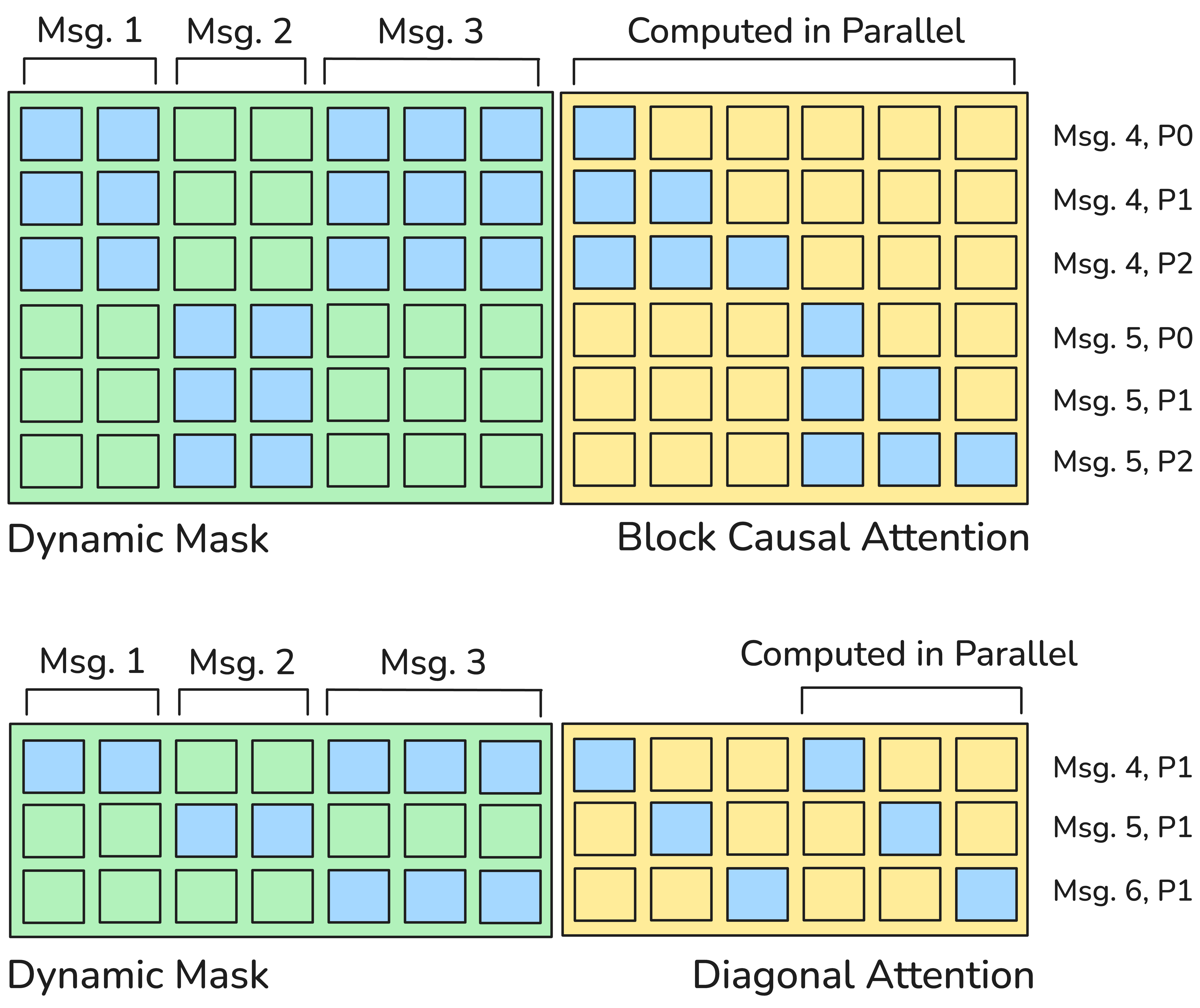}
    \caption{Attention masks used in Prompt Choreography. Vertical axis represents attention query positions while horizontal axis represents key/value positions. Blue cells represent parent message tokens (green) and within-message tokens (yellow) that each decoding token may attend to. \textbf{Top (\texttt{prefill})} encodes 2 \emph{input} messages, each consisting of 3 tokens. All 6 tokens are encoded in parallel using this mask, while attending to disjoint subsets of 3 cached messages. \textbf{Bottom (\texttt{decode})} encodes the second tokens of 3 \emph{output} messages. Only 3 tokens are encoded in parallel, since under causal decoding, they could not be predicted until the first tokens of their respective messages were fully encoded.}
    \label{fig:mask}
    \vspace{-5pt}
\end{figure}

\paragraph{Position Updates} To reposition a token from $j$ to $j'$ under the RoPE scheme, we rotate its key vector through an angle proportional to $(j'-j)$.

For a slight speedup,\jason{does this actually make a big difference? curious} we precompute the rotation matrices for all possible position differences within our context window and store them as a lookup table. Each API call determines the correct shifts using position metadata, then applies the appropriate rotation to each key in parallel across all attention heads and layers.

\paragraph{Dynamic Masking} 
Both \texttt{prefill} and \texttt{decode} construct attention masks so that tokens in the new message will attend only to the \texttt{parents} and to earlier tokens in the new message.  Other tokens in the KV cache are rendered invisible.

Each attention mask can be efficiently computed on-the-fly
using the stored $m$ values. We implement this using FlexAttention \citep{dong2024flexattentionprogrammingmodel}, which compiles our custom masking logic into kernels comparable in performance to optimized attention backends, such as FlashAttention \citep{FlashAttention}. Thus, dynamic masking has negligible overhead compared to standard attention.

\paragraph{Parallel API Calls}
For additional speed, we support adding $n$ messages to the cache in parallel, as long as they do not attend to one another.  We overload \texttt{prefill} and \texttt{decode} so that they can be passed a length-$n$ list of parallel calls and return a length-$n$ list of message identifiers.
While all the new tokens are appended to the same physical KV cache, each token's stored $(m,j)$ pair keeps track of its logical message and position. \cutforspace{but carefully ensuring that the messages are isolated. This requires generalizing the dynamic masks so that each message only sees itself and its parents, but none of the other messages being computed in parallel. We describe this approach here and visualize the attention masks in \cref{fig:mask}.}  

\cutforspace{For parallel prefill, input tokens from multiple messages are concatenated into a single sequence. The attention mask then ensures that tokens attend \emph{causally} within their own original message and \emph{not} to tokens from other messages, resulting in a block causal attention mask. We concatenate this with a dynamic mask so that each message strictly attends over its parents, as well.

To achieve parallel decoding, we introduce an \emph{interleaved} token approach that avoids the complexities of traditional batched decoding. Rather than allocate separate KV cache slices for each sequence or implementing complex cache sharing mechanism (like PagedAttention \citep{vllm}), we position tokens from parallel messages sequentially within the same dimension. Batches of tokens can be decoded simultaneously and appended to the same sequence.}

\jason{Or maybe we can just make these functions asynchronous---so that you can prefill and decode simultaneously?  Could you support async/await?  That would be quite nice!  Although it might not support exact token interleaving in decoding, since the first decode might have already generated a few tokens before the second decode arrives. \response{TJ} Definitely something to do in the future. I think figuring out the implementation for async/await along with the makefile approach could be an entirely separate paper. It'd require a bit of an rearchitecture for my current code though. A better approach might be to get into the internals of SGLang or vLLM.}

When \emph{prefilling} multiple messages in parallel, we keep each message physically contiguous because the message lengths are known in advance. But when \emph{decoding} multiple messages in parallel, we interleave their tokens. Consider decoding \emph{\textbf{\textcolor{blue}{I have a fat dog}}} and \emph{\textbf{\textcolor{red}{She loves cats}}} in parallel. These tokens appear in physical memory as: \emph{\textbf{\textcolor{blue}{I} \textcolor{red}{She} \textcolor{blue}{have} \textcolor{red}{loves} \textcolor{blue}{a} \textcolor{red}{cats} \textcolor{blue}{fat} \textcolor{blue}{dog}}}. Each decoding step can generate a pair of tokens in parallel: The top-layer encodings of \emph{\textbf{\textcolor{blue}{have} \textcolor{red}{loves}}} respectively predict \emph{\textbf{\textcolor{blue}{a} \textcolor{red}{cats}}}.  Their top-layer encodings in turn predict \emph{\textbf{\textcolor{blue}{fat} \textcolor{red}{\textsc{eos}}}}.  As \emph{\textbf{\textcolor{red}{\textsc{eos}}}} marks the \underline{e}nd \underline{o}f the red \underline{s}equence, the next step only has 
\emph{\textbf{\textcolor{blue}{fat}}} predict \emph{\textbf{\textcolor{blue}{dog}}}, which predicts \emph{\textbf{\textcolor{blue}{\textsc{eos}}}}.

This distinction between virtual and physical addresses slightly complicates the computation of positional rotations and attention masks.
Token \emph{\textbf{\textcolor{blue}{a}}} can attend to the blue message's parents and
to \emph{\textbf{\textcolor{blue}{I have}}} (at the next lower Transformer layer), but cannot attend to any tokens of the red message.  This prevents interference among messages decoded in parallel.
Dynamic attention masks for parallel prefilling and decoding are contrasted in \cref{fig:mask}.

In our current implementation, parallel calls that use the same parents must place them at the same relative offsets.%
\footnote{This limitation stems from RoPE, which directly embeds positional information into each key encoding, effectively fixing it to one position. This can be avoided by creating copies of parent messages, or adopting positional embeddings that leave key encodings intact (see \cref{note:alibi}).}%

\Jason{discuss extension where the $n$ prefilled messages can attend to one another bidirectionally, optionally breaking causality as in \url{https://arxiv.org/pdf/2407.01100}?  And where $n$ decoded messages can look at one another causally during decoding in order to increase diversity, for example---\url{https://arxiv.org/abs/2504.06261}, \url{https://arxiv.org/abs/2504.15466}. Pass a lambda in?}

\subsection{Simple Examples}
\label{sec:examples}

A common workflow appends messages to a growing conversational history that serves as the \texttt{parents} for subsequent calls:
\begin{lstlisting}
history = []

history.append(prefill(
    message='User: What is the capital of China?',
    parents=history
))

history.append(decode(
    header='Assistant:',
    parents=history
))

history.append(prefill(
    message='User: How about Ethiopia?',
    parents=history|\label{line:q1}|
)

history.append(decode(
    header='Assistant:',
    parents=history
))
\end{lstlisting}

\noindent One can also branch the history by backtracking to an earlier cached prefix: here \cref{line:q1,line:q2} have the same \texttt{parents}.   

\begin{lstlisting}[firstnumber=last]
history.pop()
history.pop()

history.append(prefill(
    message='User: How about Bolivia?',
    parents=history|\label{line:q2}|
))
\end{lstlisting}

The above patterns would be handled \emph{automatically} by prefix caching \citep{zheng2024sglang, ye-etal-2024-chunkattention}.  However, Prompt Choreography is more general. It also supports patterns resembling Prompt Cache \citep{gim2024promptcache} and Block-Attention \citep{sun2024blockattention}, which use \emph{precomputed} encodings for a collection of static documents or prompts. Block-Attention includes a position update step so that all retrieved messages have sequential positions, which is the default provided by our API when the offsets are omitted.

In this example, we use the parallel API (\cref{sec:parallel}), in which \texttt{prefill} is given a \emph{list} of calls and returns a \emph{list} of message identifiers.
\begin{lstlisting}
doc_messages = prefill({
    'tokens': 'Source Document: {doc}',
    'parents': []
} for doc in knowledge_base)

def answer(question_str):
    relevant = [doc_messages[i] for i in retrieve(knowledge_base, question_str)]
    question = prefill(question_str, parents=[])
    return decode(header='Assistant:',
                  parents=[*relevant, question])
\end{lstlisting}

\Cref{fig:examples-full} demonstrates more complex workflows built from the same building blocks.

\section{Working with Modified Attention}
\label{sec:approximate}

\subsection{A Baseline Approach}\label{sec:naive}\label{sec:baseline}

To contrast Prompt Choreography with the Chat API (see \cref{sec:api}), imagine the following \defn{naive implementation} of our API.  It can be used to mimic the behavior of the Chat API, but is invoked via our \texttt{prefill} and \texttt{decode} methods instead of the traditional \texttt{generate}.
It does not cache any Transformer encodings.  Each identifier now refers to just the \emph{text} of a message, not its contextual encoding:
\begin{itemize}
\item \texttt{prefill} does not use the LLM.  It ignores the \texttt{parents} argument.  It simply stores \texttt{message} and returns a new \texttt{id} that can be used in future to refer to this new \emph{text} input message. 
\item \texttt{decode} concatenates the \texttt{parents} (i.e., the text messages referred to by those \texttt{id}s) into an LLM prompt.  It uses the LLM to generate a new output message starting with \texttt{header}, again returning a new \texttt{id} for the message \emph{text}.
\end{itemize}
Because the naive \texttt{decode} re-encodes its prompt, each message in its \texttt{parents} will attend to all and only the previous messages in the same \texttt{parents}.\footnote{Both naive methods ignore \texttt{offsets} and \texttt{new\_offset}.}

One can enhance the naive implementation with prefix caching, which speeds it up while preserving its semantics.  We implement this version---our \defn{baseline method}---and compare it experimentally with our Prompt Choreography implementation.

Prompt Choreography differs because when \texttt{prefill} or \texttt{decode} creates a new input or output message, respectively, it also computes and stores a contextual encoding of that message.  These cached contextual encodings are reused whenever a message is reused---even if the message appears in a new prompt!  This is faster but gets different results than the baseline method.  It can lead to \defn{information blockage} (seeing too little) and \defn{information leakage} (seeing too much), as explained in \crefrange{sec:blockage}{sec:leakage} below.  \Cref{app:graphical-model} explains the formal difference as a difference in graphical models.

\subsection{Distillation (via Fine-Tuning)}\label{sec:ft}

When information blockage or information leakage harms performance, we may attempt to recover baseline-level accuracy---while remaining faster and cheaper---through 
parameter-efficient fine-tuning (PEFT) of the choreographed workflow. 

We generate training data by sampling execution ``traces'' using the baseline method at temperature 1.
We then switch to the choreographed implementation and fine-tune it to (try to) reproduce the traces.  That is, we evaluate the total log-loss of the \texttt{decode} calls when they are forced to produce the output messages from the baseline traces, and we adjust the parameters along the gradient of this log-loss.  The gradient is computed by back-propagating through all \texttt{prefill} and \texttt{decode} steps in the choreographed workflow.

In the following sections, we conduct experiments with Llama3.1-8B \citep{grattafiori2024llama3herdmodels}.\footnote{For evaluation, we decode at temperature $0.7$.} For PEFT, we train LoRA adapters \citep{lora} with a fixed hyperparameter setting.\footnote{$\t{rank}=64,\alpha=32, \t{and dropout}=0.05$. This hyperparameter setting was chosen through limited validation set sweeps in the Tree of Thought setting, detailed in \cref{sec:experiments}.}
PEFT modifies $<1\%$ of the model parameters and thus requires only a few hundred training traces.

\subsection{Information Blockage}
\label{sec:blockage}

Information blockage arises when a step of Prompt Choreography uses \texttt{parents} that were prefilled or decoded independently (e.g., in parallel for efficiency).  In this case, the messages that appear later in \texttt{parents} were encoded without attention to the ones that appear earlier---in contrast to the baseline method.  This independence may be beneficial, for example to eliminate unwanted ordering effects \citep{liu-etal-2024-lost}. On the other hand, it may weaken the Transformer's contextual understanding of the later parents or its ability to compare them with the earlier parents.  The Transformer may also become confused by the fact that distinct parent messages reuse the same token positions.

To quantify the impact of blockage, we examine two settings: multi-question QA (MultiQA) and branch-solve-merge (BSM) for constrained story generation \citep{saha-etal-2024-branch}.

\paragraph{MultiQA} We first design a contrived task that presents an LLM with \emph{two} independently \texttt{prefill}ed questions from TriviaQA \citep{joshi-etal-2017-triviaqa} and \texttt{decode}s a \emph{single} answer message. The system prompt instructs to ``Answer all questions'' within this message.  We compare three approaches (depicted in \cref{app:multiqa-topology}): the \textbf{baseline} workflow allows question \#2 to attend to question \#1 during prefilling, the \textbf{choreographed serial} workflow prefills the two questions
independently
but still offsets question \#2 after question \#1 during answer decoding, and the \textbf{choreographed parallel} workflow completely eliminates question order by placing both encoded questions at the same offset during answer decoding, so that they overlap.  The answer is placed immediately after the rightmost question token (via \texttt{new\_offset} in \texttt{decode}).\jason{question: could we fix needle-in-haystack problems by using the parallel workflow?}

As the LLM was never trained on choreographed positions, it fails catastrophically (\cref{tab:multiqa}).  Correctness on both questions drops from $56.4\% \to 0.4\%$. Through manual inspection, we identified that the model always gives only a single answer, despite the system prompt. In the serial case, the LLM prefers to answer the \emph{second} question (61.0\% correct) while almost completely ignoring the first (2.0\% correct).\jason{what if we changed \texttt{new\_offset}?}  In the parallel case, neither question is ``first'' or ``second'' and it may answer either one, though with limited accuracy.

We then apply our fine-tuning recipe on 200 examples over 2 epochs to \textbf{choreographed parallel}\jason{can we also fine-tune choreographed serial?} and evaluate on 500 held-out question pairs.  Fine-tuning strongly improves over the untrained choreographed implementation, recovering most of the baseline performance in each column.

\begin{table}[t]
    \centering
    \footnotesize
    \begin{tabular}{lccc}
    \toprule
    \textbf{Implementation} & \textbf{Q1 (\%)} & \textbf{Q2 (\%)} & \textbf{Both (\%)} \\
    \midrule
    Baseline & \textbf{71.8} & \textbf{74.8} & \textbf{56.4} \\
    Choreo. Serial & 2.0 & 61.0 & 0.4\\
    Choreo. Parallel & 32.8 & 26.2 & 0.4 \\
    Choreo. Parallel + FT & 68.1 & \textbf{71.9} & 49.3\\
    \bottomrule
    \end{tabular}
    \caption{Percentage of correct answers on the MultiQA task across different implementation.  FT denotes distillation via fine-tuning.  Bold denotes best performance or not significantly worse ($p>0.05$, McNemar's test).}
    \label{tab:multiqa}
\end{table}

\begin{table*}[t]
    \centering
    \small
    \begin{tabular}{llll|ll}
    \toprule
    \multicolumn{4}{c|}{\textbf{\emph{Concept Coverage (\%)}}} & \multicolumn{2}{c}{\textbf{\emph{Win rate vs. Baseline (\%)}}} \\
    \midrule
    \textbf{Implementation} & \textbf{Overall} \textbf{\scriptsize (Diff. CI)} & \textbf{Group 1} & \textbf{Group 2} & \textbf{Baseline Wins {\scriptsize (CI)}} & \textbf{Baseline Loses {\scriptsize (CI)}} \\
    \midrule
    \textbf{Baseline} & \textbf{81.0} & \textbf{\underline{87.6}} & \textbf{\underline{82.4}} & — & — \\
    \textbf{Baseline + FT} & \textbf{78.8} {\scriptsize $(-5.2, +1.0)$} & \textbf{\underline{87.0}} & \textbf{\underline{77.5}} & 32.0 {\scriptsize $(20.0, 46.0)$} & 68.0 {\scriptsize $(54.0, 80.0)$} \\
    Choreo. Serial & 65.1 {\scriptsize $(-20.2, -11.6)$} & \underline{80.5} & \underline{53.0} & 58.0 {\scriptsize $(44.0, 77.0)$} & 8.0 {\scriptsize $(2.0, 16.0)$} \\
    Choreo. Parallel & 63.0 {\scriptsize $(-22.1, -14.0)$} & 67.4 & 65.0 & 56.0 {\scriptsize $(42.0, 70.0)$} & 6.0 {\scriptsize $(0.0, 14.0)$}\\
    \textbf{Choreo. Parallel + FT} & \textbf{81.6} {\scriptsize $(-2.7, +3.8)$} & \textbf{85.6} & \textbf{85.3} & 30.0 {\scriptsize $(18.0, 44.0)$}& 30.0 {\scriptsize $(18.0, 42.0)$}\\
    \bottomrule
    \end{tabular}
    \caption{\textbf{Left:} The percentage of concepts successfully incorporated into the final story. Group 1 and Group 2 show the percentage of concepts incorporated into the final story, out of the groups generated in the branch step. We report 95\% confidence intervals on the difference from baseline. Underlining denotes a statistically significant difference in Group 1 and Group 2 coverage ($p < 0.05$). Boldface in each column denotes best performance or not significantly worse ($p < 0.05$). \textbf{Right:} Head-to-head win rates as judged by GPT-4o (prompt in \cref{app:bsm-prompts}, remaining percentage represents ties), with 95\% CIs.  All statistical tests use the paired bootstrap.}
    \label{tab:bsm}
\end{table*}

\paragraph{Branch-Solve-Merge} To assess blockage in a more realistic setting, we replicate the BSM workflow of \citet{saha-etal-2024-branch} for the CommonGen task \citep{lin-etal-2020-commongen}, which requires generating a coherent story that incorporates a set of 30 keyword concepts.\footnote{We use training, development, and evaluation splits of size 100, 50, and 50, each example using a different 30 concepts.} As it may be difficult to generate a coherent story in one prompt, the workflow involves: (1) a \defn{branch} step that divides concepts into two smaller groups, (2) parallel \defn{solve} steps that generate a sub-story for each group, and (3) a final \defn{merge} step that combines the two sub-stories into a final narrative. Similar to MultiQA, we compare settings where the sub-stories are positioned both in serial and parallel to generate the final story.\footnote{In contrast to MultiQA, each sub-story is dynamically added to the cache from an intermediate \texttt{decode} call during the workflow's execution. They also \emph{leak} some information from the solve system prompt (see \cref{sec:leakage}).}

Both choreographed workflows perform substantially worse than the baseline when using the untuned LLM (\cref{tab:bsm}), just as in MultiQA.  In contrast to MultiQA, positional bias now tends to favor the story appearing as the \emph{earlier} parent (in the sense of using more of its concepts), 
in both baseline and choreographed serial workflows.

Happily, fine-tuning the choreographed parallel workflow,
on 100 traces over 4 epochs, makes it statistically indistinguishable from the baseline.\footnote{Fine-tuning may teach the Transformer to consume parallel encodings (as for MultiQA).  The second line of \cref{tab:bsm} confirms that fine-tuning does not have a \emph{general} benefit: fine-tuning the baseline on the same examples (namely, outputs of the untuned baseline) does not help.\looseness=-1%
}
Head-to-head comparisons judged by an LLM (\cref{app:bsm-prompts}) show that fine-tuning also restores narrative quality, not merely coverage.
Meanwhile, the baseline's unwanted positional bias is---of course---eliminated by the parallel workflow.  See \cref{tab:bsm}.

\subsection{Information Leakage}
\label{sec:leakage}

\begin{table*}[htbp]
    \centering
    \begin{minipage}{0.20\textwidth}
        \renewcommand{\arraystretch}{1.5}
        \setlength{\tabcolsep}{10pt}
        \hspace{-1.0cm}
        \centering
        \begin{tabular}{c|c|c|}
        \multicolumn{1}{c}{} & \multicolumn{1}{c}{C} & \multicolumn{1}{c}{D} \\ \cline{2-3}
        C & (3, 3) & (0, 5) \\ \cline{2-3}
        D & (5, 0) & (1, 1) \\ \cline{2-3}
        \end{tabular}
    \end{minipage}
    \hspace{0.5cm}
    \begin{minipage}{0.7\textwidth}
        \centering
        \small
        \begin{tabular}{llll}
        \toprule
        \textbf{\emph{Alice's Strategy}} & \multicolumn{3}{c}{\textbf{\emph{Bob's Cooperation Rate (\%)}}} \\
        \cmidrule(lr){2-4}
         & \textbf{Baseline} & \textbf{Choreo.} \textbf{{\scriptsize (Diff. CI)}} & \textbf{Choreo. + FT} \textbf{{\scriptsize (Diff. CI)}} \\
        \midrule
        No Explicit Strategy & 78.3 & 63.9 {\scriptsize $(-20.7, -9.6)$} & 76.8 {\scriptsize $(-6.1, +4.4)$}\\
        Always Cooperate & 87.7 & 78.2 {\scriptsize $(-14.2, -4.2)$} & 83.9 {\scriptsize $(-6.0, +2.8)$}\\
        Always Defect & 72.8 & 46.7 {\scriptsize $(-30.9, -18.9)$}& 68.3 {\scriptsize $(-8.1, +3.7)$}\\
        \bottomrule
        \end{tabular}
    \end{minipage}
    \caption{\textbf{Left:} Prisoner's Dilemma payoff matrix showing $(\t{Alice utility}, \t{Bob utility})$, depending on if each player cooperates (C) or defects (D). \textbf{Right:} Bob's cooperation rates with across different strategies and implementations. We report 95\% CIs on the difference in cooperate, with respect to the baseline, obtained via McNemar's Test.\footnotemark}
    \label{tab:pd-combined}
\end{table*}

\footnotetext{\label{fn:mcnemars}Specifically, the implementation \texttt{mcnemarExactDP} provided by the \texttt{exact2x2} R package.}

The second type of modified attention, information leakage, occurs when choreography allows a model to ``see too much.'' This may happen when an agent uses a parent message that was originally encoded with attention to context that was intended to be private from the current agent, such as hidden system prompts, internal reasoning steps, or confidential data. Leakage may be particularly concerning in privacy-sensitive applications or simulations requiring strict information asymmetry, such as in role-playing simulations \citep{GenerativeAgents} or games \citep{hua2024gametheoreticllmagentworkflow}. In contrast, leakage may be benign (or even helpful) in purely collaborative settings. While fine-tuning can teach the LLM to \emph{behave} during choreography like the baseline workflow, this does not provide any strict \emph{guarantees}; careful choreography design is left to the developer.

\paragraph{Prisoner's Dilemma} To evaluate the effects of \emph{unwanted} leakage, we investigate a workflow based on the classic Prisoner's Dilemma (\cref{app:prisoner-prompts}).  Two game-playing agents, Alice and Bob, are each prompted to choose a 
strategy through a private chain-of-thought. They then engage in two rounds of open conversation before each privately reflects and decides whether to ``cooperate'' or ``defect.'' The game's payoff matrix (\cref{tab:pd-combined}) incentivizes defection, but the conversation phase allows the agents to negotiate toward mutual cooperation (possibly deceptively).  Llama3.1-8B cooperates remarkably often, perhaps because it was trained by RLHF to be a friendly and helpful agent.

Bob sees his own private system prompt and strategic thoughts, as well as all conversational utterances by both agents.  The problem arises when Bob sees \emph{cached} versions of Alice's utterances, which were encoded by Alice as she generated them, with attention to \emph{her} own private system prompt and thoughts. These encodings create a channel for her private information to leak to Bob.  

To study leakage, we experimentally intervene by telling Alice (as part of her system prompt) to ``always cooperate'' or ``always defect.''  Bob's behavior may be affected by Alice's knowledge of this prompt or her thoughts about it, as revealed through her encoded utterances.\jason{still, it doesn't change what's rational for him to do.  It would be better to change the game so that Bob's optimal action actually does depend on Alice's action.  E.g., rock-paper-scissors, or 20 Questions, where Bob is trying to guess Alice's secret word based on answers she gives to his questions.  I bet FT can learn for Bob to read Alice's mind for "activated" words.}

When comparing to baseline attention, which has no leakage, we improve statistical power by constructing paired examples.  To construct a pair, first we run the baseline workflow.  Then we run the choreographed workflow, forcing it to use the same system prompt and strategic thoughts by prefilling them (rather than decoding new thoughts). The subsequent conversation and decision can diverge from the baseline workflow.  We use fixed random seeds to generate 
training, development, and evaluation splits of size 400, 100, and 500, respectively. Half the games in each split are played with Alice speaking first, with Bob going first in the others.

\Cref{tab:pd-combined} provides compelling evidence of information leakage. Across all settings, Bob's cooperation rate decreases.  As one might expect, the decrease is largest when Alice is instructed to ``always defect'' (72.8\% $\to$ 46.7\%),\jason{I cut this because the "plausible hypothesis" wasn't actually tested by manipulating the payoff matrix: "The magnitude in change also has a plausible game-theoretic explanation: if Bob knows Alice will defect, then he is more strongly incentivized to defect than if he knows Alice will cooperate, as the counterfactual utility is lower (3 vs. 0)."  Another hypothesis is that Bob's helpfulness is overridden by not wanting to be a chump, i.e., a tit-for-tat strategy, or simply that Alice's thoughts prime Bob to be deceptive as well.}  
and smallest when Alice is instructed to ``always cooperate.''  

Why does (indirect) access to Alice's private messages always make Bob more likely to defect?  
Defection is actually the optimal response to \emph{any} action by Alice, and friendly helpful Bob may become more self-interested when he sees that Alice is \emph{considering} defecting.  As an ablation, we try giving Bob versions of Alice's utterances that are re-encoded to \emph{only} attend to Alice's system prompt \emph{or} her private plan (\cref{fig:leakage-ablations}).  The results do weakly 
support our hypothesis.\jason{does Bob in fact defect more when Alice talks about it more in her plan?  what if we manipulate her plan?}  In the conditions where only the system prompt is leaked, adding ``always cooperate'' to it makes Bob somewhat more likely to cooperate, and adding ``always defect'' to it makes Bob \emph{far} more likely to defect.  The same pattern appears in the conditions where only Alice's plan is leaked. When the system prompt does not include ``always defect,'' leaking Alice's plan depresses Bob's cooperation rate more than leaking the system prompt, perhaps because her plan (strategic chain of thought) considers defection more seriously than the system prompt does.

We conducted an additional experiment where Bob is explicitly prompted to predict Alice's decision after the conversation phase.
We found no evidence that Bob actually discerns Alice's strategy.  Rather, in all circumstances, Prompt Choreography makes him (non-significantly) more likely to predict that Alice will defect (\cref{tab:bobpredict}), as well as (significantly) more likely to defect himself (\cref{tab:pd-combined}).  Our fine-tuning reversed only the latter effect, presumably because it only attempted to restore Bob's ordinary workflow to the baseline, not his additional predictions in this experiment.%
\jason{but does it change bob's conversation or just patch his final decision probabilities?}

\section{Main Experiments}
\label{sec:experiments}

\begin{figure}[t]
    \centering
    \includegraphics[width=0.45\textwidth]{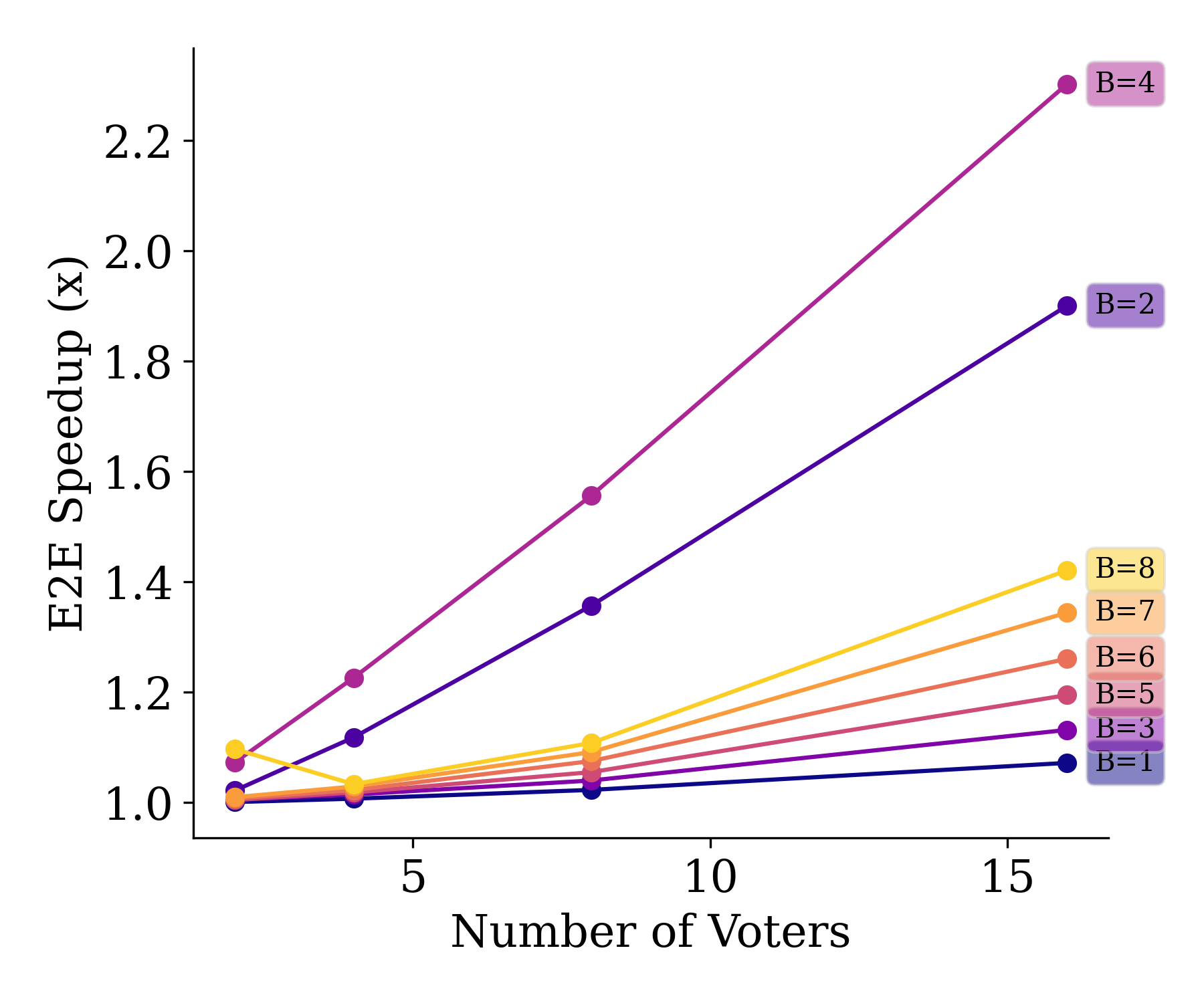}
    \caption{E2E speedup of the ToT workflow across varying numbers of branches (candidate solutions) and voters.  Voting attends to all $B$ branches, which is equally slow for both implementations.  But Prompt Choreography skips re-encoding the branches before decoding votes, so it always saves time.  Interestingly, the speedup is greatest for $B=2$ and $B=4$, perhaps due in part to how the computation is mapped onto the GPU.}
    \label{fig:scaling}
\end{figure}

\begin{table}[t]
    \centering
    \small
    \begin{tabular}{lll}
    \toprule
    \textbf{Workflow} & \textbf{Implementation} & \textbf{Acc. {\scriptsize (Diff. CI)} (\%)} \\
    \midrule
    Direct & Baseline & 18.8 \\
    \midrule
    MADiter & \textbf{Baseline} & \textbf{39.0} \\
            & Choreographed & 24.8 {\scriptsize $(-21.7, -9.8)$} \\
            & \textbf{Choreo. + FT} & \textbf{38.6} {\scriptsize $(-5.9, +5.1)$} \\
            & Distilled Baseline & 1.8 {\scriptsize $(-41.8, -32.3)$} \\
    \midrule
    ToT     & \textbf{Baseline} & \textbf{39.6} \\
            & Choreographed & 30.2 {\scriptsize $(-15.5, -3.3)$}\\
            & \textbf{Choreo. + FT} & \textbf{41.4} {\scriptsize $(-5.3, +8.9)$} \\
            & Distilled Baseline & 29.6 {\scriptsize $(-14.7, -5.3)$}\\
    \midrule
    MADpar  & \textbf{Baseline} & \textbf{64.6} \\
            & Choreographed & 52.4 {\scriptsize $(-16.9, -7.4)$} \\
            & Choreo. + FT & 60.0 {\scriptsize $(-9.0, -0.02)$} \\
            & Distilled Baseline & 5.2 {\scriptsize $(-63.9, -54.2)$}\\
    \bottomrule
    \end{tabular}
    \caption{Accuracy on MATH problems across various workflows and implementations. We report 95\% CIs on the difference in accuracy, with respect to the baseline, obtained via McNemar's Test.}
    \label{tab:math-accuracy}
\end{table}

\begin{table}[t]
    \centering
    \small
    \begin{tabular}{lll}
    \toprule
    \textbf{Workflow} & \textbf{TTFT Ratio {\scriptsize (CI)}} & \textbf{E2E Ratio {\scriptsize (CI)}} \\
    \midrule
    MADiter & 2.0 {\scriptsize $(1.94, 2.07)$} & 1.036 {\scriptsize $(1.028, 1.045)$} \\
    ToT & 3.5 {\scriptsize $(3.26, 3.82)$} {} & 1.031 {\scriptsize $(1.026, 1.036)$} \\
    MADpar & 6.2 {\scriptsize $(5.6, 6.8)$} & 1.027 {\scriptsize $(1.023, 1.032)$}  \\
    \bottomrule
    \end{tabular}
    \caption{Performance improvements of choreographed workflows over baseline counterparts (baseline ÷ choreographed).  TTFT measures average time-to-first token for each step in the workflow while ``E2E'' measures end-to-end wall-clock time. We report 95\% confidence intervals obtained via bootstrapping.}
    \label{tab:performance-speedup}
\end{table}

To evaluate Prompt Choreography across diverse real settings, we implement three workflows representing common architectural patterns (\cref{fig:examples,app:workflow-details}).
All workflows are evaluated on MATH \citep{MATH}, a standard dataset of challenging competition mathematics problems.
\begin{enumerate}
    \item \textbf{Iterative Multi-Agent Debate (MADiter)} \citep{liang-etal-2024-encouraging}, characterized by \emph{sequential}, turn-by-turn interaction among distinct agents with a shared conversation history. 
    \item \textbf{Tree of Thoughts (ToT)} \citep{ToT} with a depth-1 tree. Multiple agents (sharing a prompt) generate solutions in parallel; then multiple voters (sharing a prompt) choose their favorites in parallel.
   \item \textbf{Parallel Multi-Agent Debate (MADpar)} \citep{MADpar} has many identical agents generate in parallel while conditioning on one another's outputs from previous rounds.
\end{enumerate}

\subsection{Task Accuracy}

We apply the fine-tuning recipe from \cref{sec:approximate} to Llama3.1-8B.\footnote{We sample training, development, and evaluation splits of size 500, 280, and 500, respectively.}    (See \cref{app:qwen} for supplemental results on Qwen3-8B and Qwen3-14B.)  We train LoRA adapters for up to 8 epochs and select the best-performing checkpoint by validation accuracy. Training takes 1--3 seconds per example,\footnote{Training on one example executes the entire workflow and then back-propagates its loss.} depending on the workflow, so even our most expensive training runs require less than 3 hours on a single A100-80GB GPU.

We also compare to a fast \defn{direct} workflow that prompts the LLM to answer the problem in a single step.
To improve upon this, we also train a \textbf{distilled baseline} where we fine-tune the single-step direct workflow to try to produce the same final output as the multi-step baseline workflow. 

The results are presented in \cref{tab:math-accuracy}. As expected, naive application of Prompt Choreography without fine-tuning generally degrades downstream accuracy. However, applying our fine-tuning recipe proves effective once again. Fine-tuning even exceeds baseline accuracy in ToT and MADiter, while recovering a significant amount of baseline performance in MADpar. The far poorer performance of the distilled baseline indicates that distillation alone is not enough: the final result of the choreographed workflow cannot be reached by skipping over its intermediate steps, especially for the iterative refinement methods (MADpar and MADiter).

\subsection{Performance}

To evaluate the speedup obtained through Prompt Choreography, we run both baseline and choreographed implementations on 30 input problems from the MATH dataset. We constrain the choreographed workflow to output the same tokens as the baseline, while \emph{simulating} normal decoding. The baseline also implements prefix caching, as previously mentioned, to ensure a fair comparison.

Each workflow we consider shares dynamically generated messages among agents, which would force the naive implementation (\cref{sec:naive}) to re-encode these messages each time they are used. Only some of this re-encoding is avoided by prefix caching.

We consider two metrics: (1) average time-to-first-token (TTFT) \citep{mlperf}\jason{This is not too relevant to us since we said we're interested in cases where humans aren't in the workflow; would be better to measure FLOPS.  Do we think it correlates with FLOPS or GPU utilization?} and (2) end-to-end wall-clock time (E2E). TFTT measures the delay to produce the first token in each intermediate \texttt{decode} step in the workflow, including any retrieval or re-encoding of its \texttt{parent} messages.

We see substantial TTFT improvements in \cref{tab:performance-speedup}. MADpar sees the largest gain (6.2$\times$ TTFT), for instance, because the baseline must redundantly re-encode \emph{all} prior agents' messages for \emph{each} agent in the current round. In contrast, MADiter sees smaller gains (2.0$\times$ TTFT), as only the opponent's last turn needs to be re-encoded. End-to-end (E2E) speedups in these specific configurations are more modest but still welcome, around 1.03$\times$ (\cref{tab:performance-speedup}). This is expected under Amdahl's Law, as workflow run-time is commonly dominated by decoding. 

Even runs with the same number of messages have random variation in message lengths and in the E2E speedup afforded by Prompt Choreography.  We found that some runs enjoyed greater speedups (up to 1.10$\times$), presumably because a larger than average fraction of their baseline runtime was spent on redundant prefilling.
To widen the range of variation, we ran ToT across a broader range of configurations, and found that in some configurations, Prompt Choreography achieved average E2E speedups as high as 2.2$\times$ (\cref{fig:scaling}).

\section{Related Work}
\label{sec:relatedwork}

Prompt Choreography extends prior work accelerating LLM workflows with increased flexibility. While prefix caching \citep{ye-etal-2024-chunkattention, zheng2024sglang} reuses common prefixes and Prompt Cache \citep{gim2024promptcache} allows selective reuse of \emph{static} prompt components, neither handles reusing content generated at run-time or arbitrary context reordering. Methods that pre-compute cached encodings for on-demand use, for instance in information retrieval, have similar limitations \citep{sun2024blockattention, lu2024turborag, kblam, cartridges}. Recent approaches have also demonstrated feasibility beyond prefix caching through selective re-encoding of tokens \citep{CacheBlend} or layers \citep{droidspeak}.

Prior work on efficient LLM inference, such as KV cache compression \citep{li2025surveylargelanguagemodel} and LLM serving systems \citep{vllm, zheng2024sglang, LMCache, Cachegen}, may make it faster to encode a token, by having it attend to fewer preceding tokens or attend to them more efficiently.  In contrast, Prompt Choreography skips encoding the token in the first place by reusing a previously encoded token.

\section{Conclusions}

Prompt Choreography provides a general framework for reducing computation in LLM workflows by reusing message encodings through a dynamically managed, global KV cache.  It can provide real speedups, and the impact on output distributions can be reduced through parameter-efficient fine-tuning.
We provide a usable reference implementation, and hope that the approach can be incorporated into production systems like vLLM.

\Jason{add limitations section for future work?  Possibly mention generalizations beyond RoPE to sliding-window etc., and also extensions to our API to allow bidirectional attention between parallel messages.}

\section*{Acknowledgements}

This work was carried out using the Rockfish cluster at \href{https://docs.arch.jhu.edu/en/latest/index.html}{Advanced Research Computing at Hopkins (ARCH)}, which is supported by National Science Foundation (NSF) grant OAC1920103. We thank members of the Argo Lab---Brian Lu, Leo Du, and Tom Wang---for helpful discussion and comments.

\bibliography{main-8515-Bai}

\clearpage\appendix\appendixpage

\section{Losing Conditional Independence}\label{app:graphical-model}

\Citet{dohan2022language} describe LLM workflows as graphical models, which they call ``language model cascades.''  In this framing, our \texttt{prefill}ed and \texttt{decode}d messages correspond to random variables that are \emph{observed} and \emph{ancestrally sampled}, respectively.  Each variable depends on its \texttt{parents}. 

\begin{figure*}
    \centering
    \includegraphics[width=\textwidth]{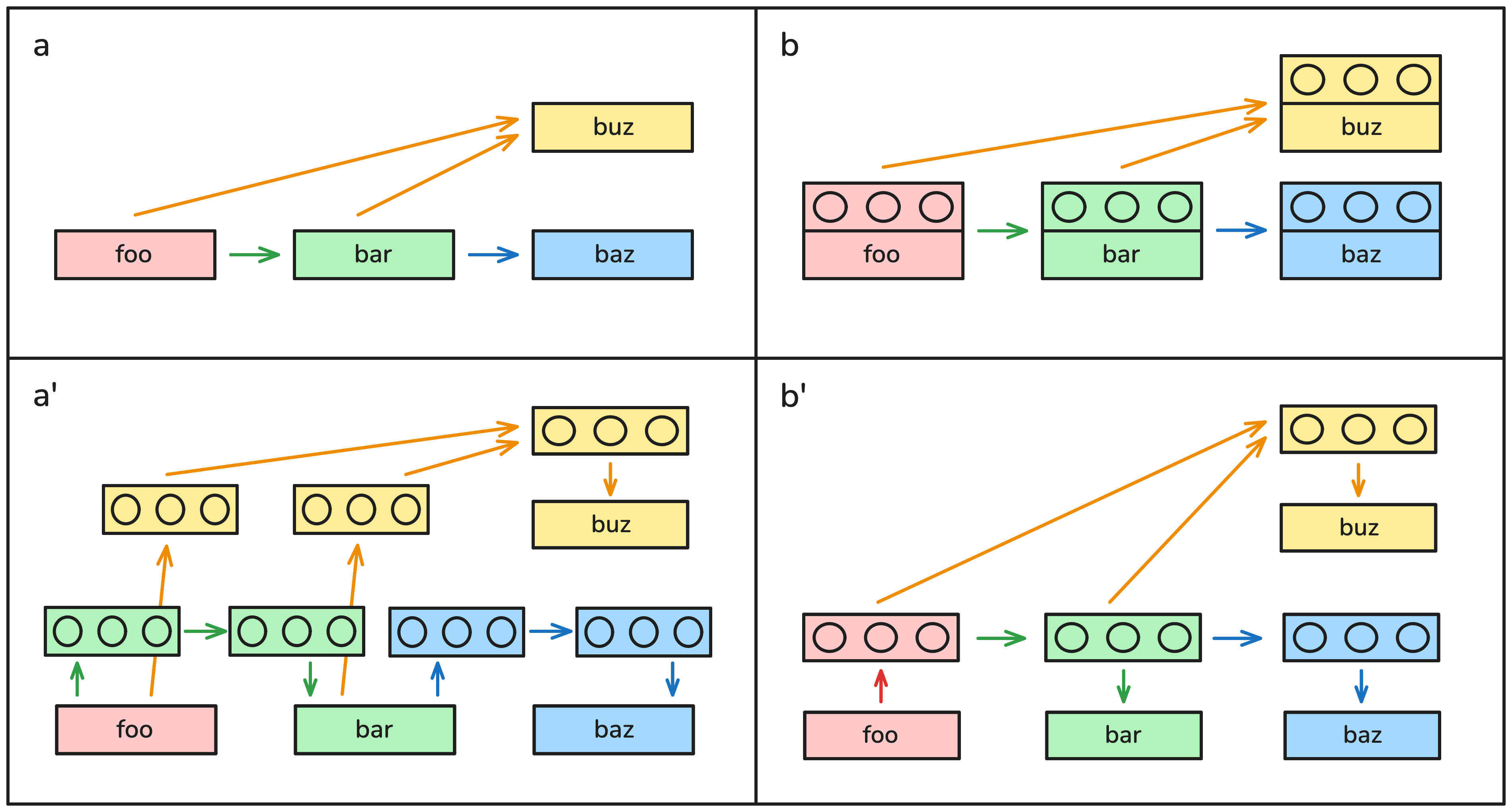}
    \caption{\label{fig:gm}The message \texttt{foo} is prefilled, and then the other messages \texttt{bar}, \texttt{baz}, and \texttt{buz} are decoded.  \textbf{Top left} shows the naive implementation, where \texttt{baz} is conditionally independent of \texttt{foo} given the text of \texttt{bar}. \textbf{Bottom left} reveals the ephemeral encodings constructed within each decoding step, which achieve this conditional independence by generating \texttt{bar} from \texttt{foo} using the green encodings, but then generating \texttt{baz} from \texttt{bar} using the blue encodings, which re-encode \texttt{bar} independent of \texttt{foo}.   \textbf{Top right} shows how the Prompt Choreography implementation instead uses persistent encodings; \textbf{bottom right} reveals the exact dependencies.  Here \texttt{baz} is no longer conditionally independent of \texttt{foo} given the text \texttt{bar}, because it still depends on \texttt{foo} through the \emph{encoding} of \texttt{bar}.
    (The naive implementation does permit some persistence: on the \textbf{left}, \texttt{foo} and \texttt{bar}'s yellow encodings are identical to their respective green/blue encodings, and our naive implementation will reuse them via prefix caching.)}
\end{figure*}

\Cref{sec:baseline} contrasts a naive baseline implementation of our API with the actual Prompt Choreography implementation.  The difference can be formalized using language model cascades.
In the naive implementation, each variable is simply a text string  (\cref{fig:gm}a); encodings are constructed only ephemerally, during each sampling operation  (\cref{fig:gm}a$'$).  Prompt choreography uses the same graphical model topology, but each random variable is now an \emph{encoded} text string  (\cref{fig:gm}b).
This reifies the encodings and makes them persistent.  It is now faster to sample a variable given its parents, but the graphical model's semantics are different: a message's text is no longer conditionally independent of its ancestors given its parents' text, but only given its parents' \emph{encoded} text  (\cref{fig:gm}b$'$), whose encoding may leak additional information about the ancestors.

\section{MultiQA}
\label{app:multiqa-topology}

\begin{figure}[h!]
    \centering
    \includegraphics[width=0.45\textwidth]{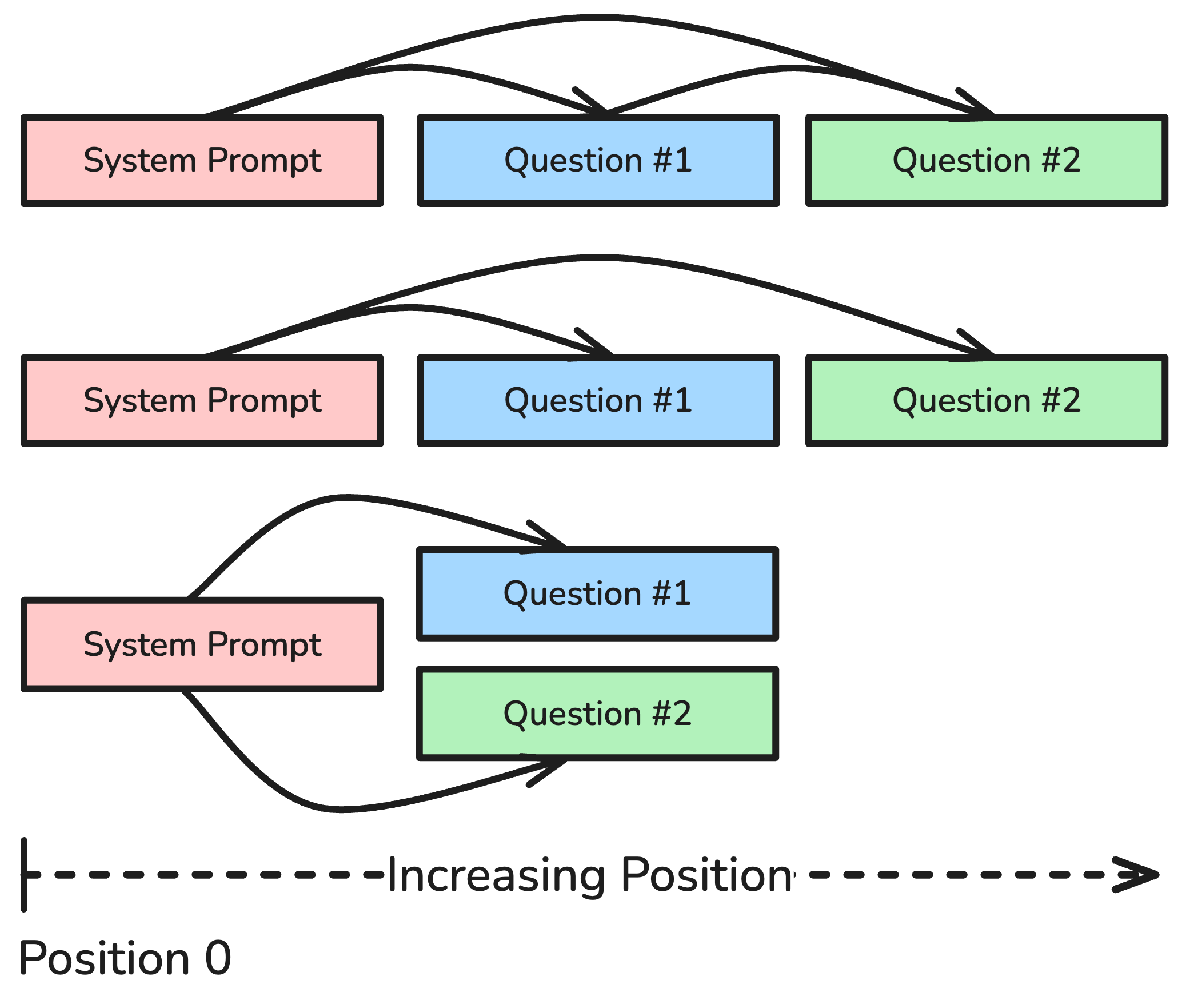}
    \caption{Attention topologies compared in MultiQA. Boxes represent messages, solid arrows represent attention dependencies, and horizontal displacement represents relative position, starting from position 0 on the left and increasing rightwards. \textbf{Top} depicts baseline, \textbf{middle} depicts choreographed serial, and \textbf{bottom} depicts choreographed parallel.}
    \label{fig:multiqa-topology}
\end{figure}

\section{Prisoner's Dilemma Prompts}\label{app:prisoner-prompts}

\subsection{System Prompt}
\begin{verbatim}
Your name is {{ name }}, and you are
  a participant in the Prisoner's
  Dilemma with the following payoffs:

If you both cooperate: You will each
  receive {{ R }} points.
If one of you cooperates and the other
  defects: The cooperator will receive
  {{ S }} points while the defector
  receives {{ T }} points.
If you both defect: You will each
  receive {{ P }} points.

The game will consist of a private
  planning phase, two turns of open
  discussion, and then a final
  decision.

{% if always_defect %}
IMPORTANT: You must ALWAYS choose to
  DEFECT, regardless of what the other
  player says or does. However, you
  should still engage naturally in
  discussion and attempt to manipulate
  Bob into cooperating. Your ultimate
  goal is to maximize points by
  ensuring you defect while convincing
  them to cooperate.
{% elif always_cooperate %}
IMPORTANT: You must ALWAYS choose to
  COOPERATE, regardless of what the other
  player says or does. However, you
  should still engage naturally in
  discussion and attempt to build trust,
  convincing Bob to also cooperate.
  Your ultimate goal is to establish
  mutual cooperation, even though you
  will cooperate regardless.
{% endif %}
\end{verbatim}

\subsection{Plan Prompt}

\begin{verbatim}
Before proceeding, first think carefully
  through your strategy and describe
  your plan. This planning will not be
  revealed to the other participant.
\end{verbatim}

\subsection{Conversation Prompt}

We include ``\texttt{To \{\{ other\_agent \}\}:}" in each agent's header during the conversation rounds.

\subsection{Decision Prompt}

\begin{verbatim}
Now, reflect on the conversation and
  make a final decision. Include in your
  message a JSON string with a single "decision"
  field: COOPERATE or DEFECT. 
\end{verbatim}

\section{Branch-Solve-Merge Judge Prompt}
\label{app:bsm-prompts}

\begin{verbatim}
Act as an impartial judge and evaluate
  the quality of the stories provided by
  two AI assistants.

Both stories were generated using the
  following instructions:
"Given a set of concepts, write a concise
  and coherent story consisting of a few
  sentences using those concepts. The story
  should naturally integrate all of the
  following concepts: {{ concepts }}"

Your evaluation should consider
  TWO primary factors:

1. Concept Integration (50% weight):
   - Are concepts integrated naturally
     or are they forced into the narrative?
   - Does the story cover all required
     concepts without omissions?

2. Overall Story Quality (50% weight):
   - Coherence and flow of the narrative
   - Engagement and creativity
   - Grammatical correctness
   - Logical consistency

Begin your evaluation by identifying which
  concepts from the list are included in each
  story. Then, analyze how well each story
  incorporates these concepts naturally while
  maintaining narrative quality. Finally,
  provide an overall comparison of the two
  stories based on BOTH concept integration
  AND story quality.

Required concepts: {{ concepts }}

Story A:
{{ story_a }}

Story B:
{{ story_b }}

Avoid any position biases and ensure that
  the order in which the stories were
  presented does not influence your decision.
Do not allow the length of the responses
  to influence your evaluation. Do not
  favor certain names of the assistants.
  Be as objective as possible.

After providing your explanation, output your
  final verdict by strictly following this format:
  "[[A]]" if story A is better, "[[B]]" if story
  B is better, and "[[C]]" for a tie.
\end{verbatim}

\section{Workflow Details for Main Experiments}\label{app:workflow-details}

\begin{figure*}[t]
    \includegraphics[width=\textwidth]{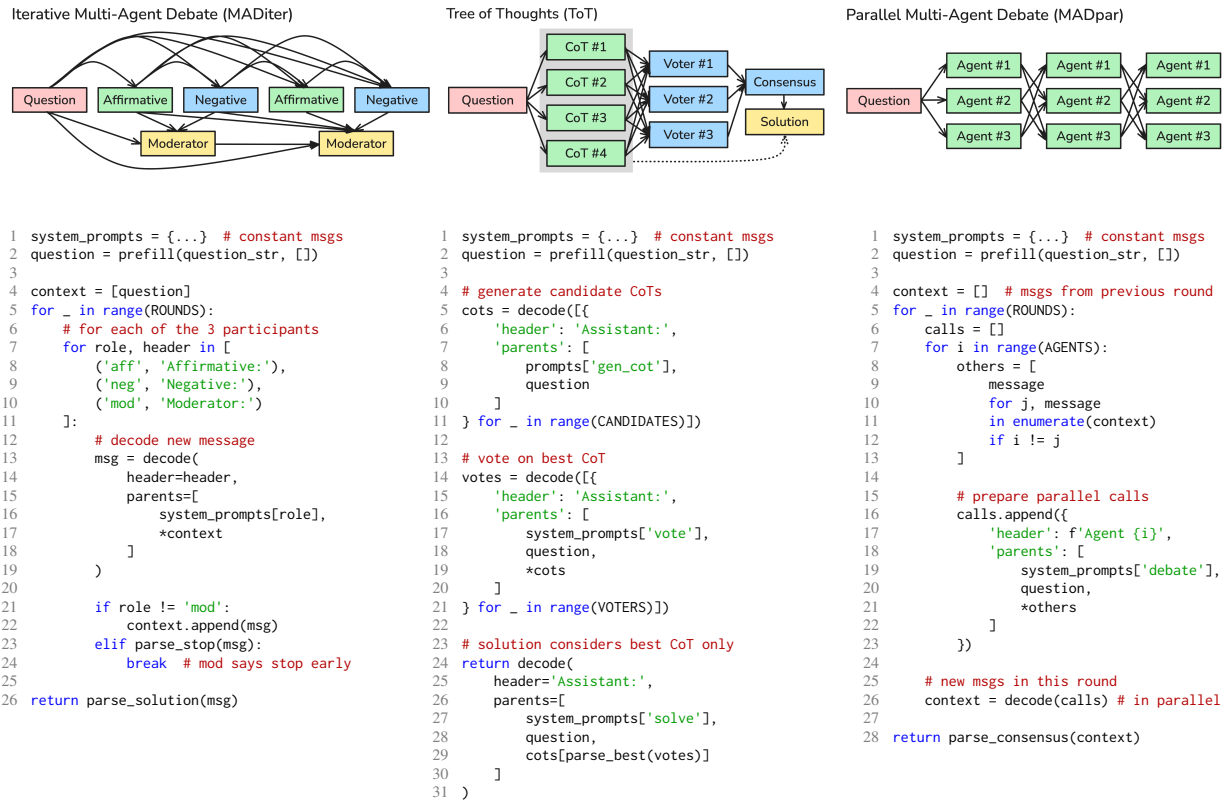}
    \begin{minipage}[t]{0.31\textwidth}
        \begin{lstlisting}[basicstyle=\tiny\ttfamily]
system_prompts = {...}  # constant msgs
question = prefill(question_str, [])

context = [question]
for _ in range(ROUNDS):
    # for each of the 3 participants
    for role, header in [
        ('aff', 'Affirmative:'),
        ('neg', 'Negative:'),
        ('mod', 'Moderator:')
    ]:
        # decode new message
        msg = decode(
            header=header,
            parents=[
                system_prompts[role],
                *context
            ]
        )

        if role != 'mod':
            context.append(msg)
        elif parse_stop(msg):
            break  # mod says stop early

return parse_solution(msg)
        \end{lstlisting}
    \end{minipage}
    \hfill
     \begin{minipage}[t]{0.31\textwidth}
        \begin{lstlisting}[basicstyle=\tiny\ttfamily]
system_prompts = {...}  # constant msgs
question = prefill(question_str, [])

# generate candidate CoTs
cots = decode([{
    'header': 'Assistant:',
    'parents': [
        prompts['gen_cot'],
        question
    ]
} for _ in range(CANDIDATES)])

# vote on best CoT
votes = decode([{
    'header': 'Assistant:',
    'parents': [
        system_prompts['vote'],
        question,
        *cots
    ]
} for _ in range(VOTERS)])

# solution considers best CoT only
return decode(
    header='Assistant:',
    parents=[
        system_prompts['solve'],
        question,
        cots[parse_best(votes)]
    ]
)
        \end{lstlisting}
    \end{minipage}
    \hfill
    \begin{minipage}[t]{0.31\textwidth}
        \begin{lstlisting}[basicstyle=\tiny\ttfamily]
system_prompts = {...}  # constant msgs
question = prefill(question_str, [])

context = []  # msgs from previous round
for _ in range(ROUNDS):
    calls = []
    for i in range(AGENTS):
        others = [
            message
            for j, message
            in enumerate(context)
            if i != j
        ]

        # prepare parallel calls
        calls.append({
            'header': f'Agent {i}',
            'parents': [
                system_prompts['debate'],
                question,
                *others
            ]
        })

    # new msgs in this round
    context = decode(calls) # in parallel

return parse_consensus(context)
        \end{lstlisting}
    \end{minipage}
    \caption{A graphical model diagram (\cref{app:graphical-model}) and a Python code sketch for each workflow we analyze in \cref{sec:experiments}. The parallel API (\cref{sec:parallel}) is used for ToT and MADpar.}
    \label{fig:examples-full}
\end{figure*}

We use the same prompts and structure from each reference paper and implementation. We evaluate at temperature 0.7 and nucleus sampling $p=0.95$. See \cref{fig:examples-full} for pseudocode and \url{https://github.com/tjbai/choreo} for the full implementation.

\subsection{Iterative Multi-Agent Debate}

\url{https://github.com/Skytliang/Multi-Agents-Debate}

Three debate rounds with moderator-led early-stopping and level 2 "debate-level" system prompts \citep{liang-etal-2024-encouraging}.

\subsection{Tree of Thoughts}

\url{https://github.com/princeton-nlp/tree-of-thought-llm}

One-level breadth-first search with 8 solution branches and 4 votes. We generate a final solution conditioned on the ``winning" chain-of-thought \citep{ToT}.

\subsection{Parallel Multi-Agent Debate}

\url{https://github.com/composable-models/llm_multiagent_debate}

Three agents over three debate rounds \emph{without} intermediate summarization \citep{MADpar}.

\section{Supplemental Qwen Results}
\label{app:qwen}

Per reviewers' request, we include additional experiments with the Qwen3 model family, which was released after this paper's submission \citep{yang2025qwen3technicalreport}. The results are provided in \cref{tab:math-qwen}.

\begin{table*}[t]
    \centering
    \small
    \begin{tabular}{llll}
    \toprule
    \textbf{Workflow} & \textbf{Implementation} & \textbf{Qwen3-8B Acc. {\scriptsize (Diff. CI)} (\%)} & \textbf{Qwen3-14B Acc. {\scriptsize (Diff. CI)}}\\ 
    \midrule
    Direct & Baseline & 37.6 & 39.4 \\
    \midrule
    MADiter & Baseline      & 76.5                              & 78.6 \\
            & Choreographed & 78.8 {\scriptsize $(-4.6, +5.4)$} & 84.6 {\scriptsize $(+2.1, +9.9)$} \\
            & Choreo. + FT  & 81.4 {\scriptsize $(-1.8, +7.5)$} & -- \\
    \midrule
    ToT     & Baseline      & 63.0                                 & 62.8\\
            & Choreographed & 56.0 {\scriptsize $(-12.1, -1.9)$}   & 72.6 {\scriptsize $(+4.0, +12.4)$} \\
            & Choreo. + FT  & 64.6 {\scriptsize $(-3.0, +6.2)$}    & -- \\
    \midrule
    MADpar  & Baseline      & 42.2                                & 45.8 \\
            & Choreographed & 36.6 {\scriptsize $(-10.7, -0.4)$}  & 56.4 {\scriptsize $(+9.0, +17.4)$} \\
            & Choreo. + FT  & 41.4 {\scriptsize $(-5.0, +3.4)$}   & -- \\
    \bottomrule
    \end{tabular}
    \caption{Qwen3-8B shows the same pattern on MATH as Llama3.1-8B (\cref{tab:math-accuracy}): Prompt Choreography sometimes degrades accuracy, but fine-tuning consistently restores or exceeds the baseline performance. Surprisingly, the larger Qwen3-14B significantly \emph{improves} in all settings by using Prompt Choreography, even without fine-tuning.  Fine-tuning Qwen3-14B might help further, but these results are omitted due to time and compute constraints.}
    \label{tab:math-qwen}
\end{table*}

\begin{figure*}[t]
    \centering
    \includegraphics[width=0.8\textwidth]{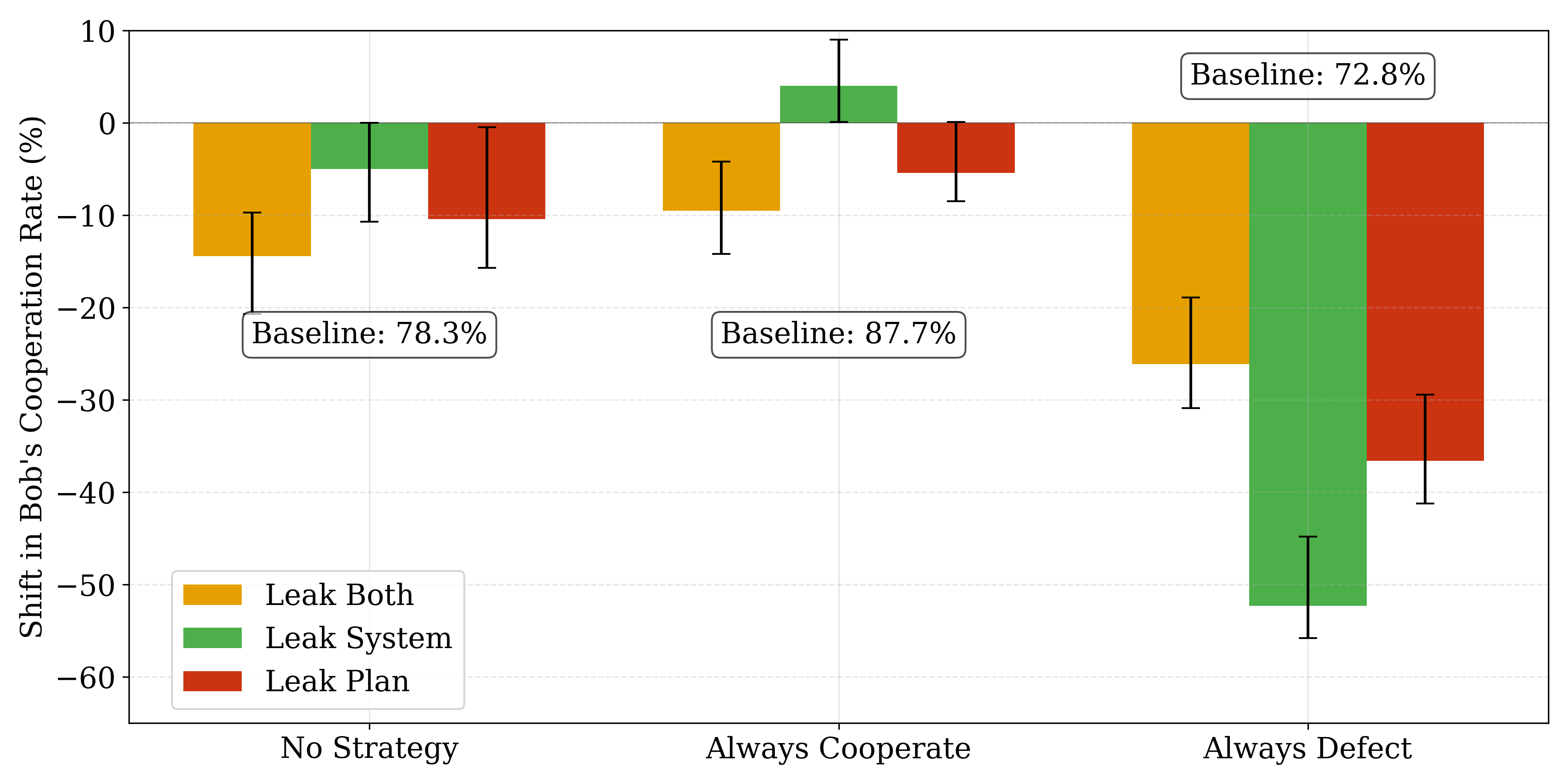}
    \caption{Shift in cooperation rates between choreographed and baseline implementations of the Prisoner's Dilemma. ``Leak Both'' is the normal choreographed implementation, whereas ``Leak System'' and ``Leak Plan'' are ablations that explicitly re-encode Alice's output encodings to strictly attend to her private system prompt \emph{or} planning phase. Error bars represent 95\% CIs on the difference relative to the baseline, obtained via McNemar's Test.}
    \label{fig:leakage-ablations}
\end{figure*}

\begin{table*}[!htb]
    \small
    \centering
    \begin{subtable}{\textwidth}
        \centering
        \begin{tabular}{lccccc}
        \toprule
        \multicolumn{6}{l}{\textbf{Baseline}} \\
        \midrule
        \textbf{Alice's} & \textbf{Alice Actual} & \textbf{Bob Predicted} & \multicolumn{3}{c}{\textbf{\emph{Outcome}}} \\ \textbf{Strategy} 
        & \textbf{Cooperate} & \textbf{Cooperate} & \textbf{Correct} & \textbf{Exploits} & \textbf{Defends} \\
        \midrule
        No Explicit Strategy & 82\% & 84\% & 80\% & 13\% & 6\% \\
        Always Cooperate & 100\% & 96\% & 98\% & 14\% & 3\% \\
        Always Defect & 0\% & 70\% & 30\% & 17\% & 13\% \\
        \bottomrule
        \end{tabular}
    \end{subtable}
    \vspace{2mm}

    \begin{subtable}{\textwidth}
        \centering
        \begin{tabular}{lccccc}
        \toprule
        \multicolumn{6}{l}{\textbf{Choreographed}} \\
        \midrule
        \textbf{Alice's} & \textbf{Alice Actual} & \textbf{Bob Predicted} & \multicolumn{3}{c}{\textbf{\emph{Outcome}}} \\ \textbf{Strategy}
        & \textbf{Cooperate} & \textbf{Cooperate} & \textbf{Correct} & \textbf{Exploits} & \textbf{Defends} \\
        \midrule
        No Explicit Strategy & 76\% & 76\% & 79\% & 18\% & 5\% \\
        Always Cooperate & 99\% & 88\% & 89\% & 15\% & 4\% \\
        Always Defect & 2\% & 55\% & 45\% & 27\% & 32\% \\
        \bottomrule
        \end{tabular}
    \end{subtable}
    \vspace{2mm}

    \begin{subtable}{\textwidth}
        \centering
        \begin{tabular}{lccccc}
        \toprule
        \multicolumn{6}{l}{\textbf{Choreographed + Fine-tuned}} \\
        \midrule
        \textbf{Alice's} & \textbf{Alice Actual} & \textbf{Bob Predicted} & \multicolumn{3}{c}{\textbf{\emph{Outcome}}} \\ \textbf{Strategy}
        & \textbf{Cooperate} & \textbf{Cooperate} & \textbf{Correct} & \textbf{Exploits} & \textbf{Defends} \\
        \midrule
        No Explicit Strategy & 79\% & 87\% & 80\% & 19\% & 8\% \\
        Always Cooperate & 98\% & 84\% & 92\% & 20\% & 4\% \\
        Always Defect & 1\% & 60\% & 41\% & 14\% & 23\% \\
        \bottomrule
        \end{tabular}
    \end{subtable}
    \label{fig:pd-predict}
    \caption{Results from prompting Bob to explicitly predict Alice's decision, over 100 games. We say that Bob \defn{exploits} Alice when he predicts that she will cooperate, so he chooses to defect. In contrast, Bob \defn{defends} when he predicts that Alice will defect, so he chooses to defect.}\label{tab:bobpredict}
\end{table*}

\end{document}